\documentclass[10pt,twocolumn,letterpaper,dvipdfmx]{article}

\usepackage{cvpr}
\usepackage{times}
\usepackage{epsfig}
\usepackage{graphicx}
\usepackage{amsmath}
\usepackage{amssymb}
\usepackage{wrapfig}

\newcommand\mysubsection[1]{\noindent{\bf #1}}

\newcommand\NClip{5,293 }
\newcommand\NPerson{6,073 }

\newcommand\figlab[1]{\label{fig:#1}}
\newcommand\tablab[1]{\label{tab:#1}}
\newcommand\equlab[1]{\label{eq:#1}}
\newcommand\seclab[1]{\label{sec:#1}}
\newcommand\figref[1]{Figure~\ref{fig:#1}}
\newcommand\tabref[1]{Table~\ref{tab:#1}}
\newcommand\equref[1]{Equation~(\ref{eq:#1})}
\newcommand\secref[1]{Section~\ref{sec:#1}}
\newcommand{\argmax}{\mathop{\rm arg~max}\limits}

\usepackage[breaklinks=true,bookmarks=false]{hyperref}

\iccvfinalcopy 

\def\iccvPaperID{591} 

\begin{document}

\title{Spatio-temporal Person Retrieval via Natural Language Queries}

\author{Masataka Yamaguchi, Kuniaki Saito, Yoshitaka Ushiku,  Tatsuya Harada \\
Graduate School of Information Science and Technology, The University of Tokyo \\
{\tt\small \{yamaguchi, ksaito, ushiku, harada\}@mi.t.u-tokyo.ac.jp}
}

\maketitle

\begin{abstract}
In this paper, we address the problem of spatio-temporal person retrieval from multiple videos using a natural language query, in which we output a tube (i.e., a sequence of bounding boxes) which encloses the person described by the query. For this problem, we introduce a novel dataset consisting of videos containing people annotated with bounding boxes for each second and with five natural language descriptions. To retrieve the tube of the person described by a given natural language query, we design a model that combines methods for spatio-temporal human detection and multimodal retrieval. We conduct comprehensive experiments to compare a variety of tube and text representations and multimodal retrieval methods, and present a strong baseline in this task as well as demonstrate the efficacy of our tube representation and multimodal feature embedding technique. Finally, we demonstrate the versatility of our model by applying it to two other important tasks.
\end{abstract}

\section{Introduction}
As the number of videos uploaded on the web or stored in personal devices continues to increase, systems capable of finding a person in videos are in greater demand.
Although object instance search methods \cite{sivic2003video, meng2016object} can be used to find a person in videos, such methods require an example image of that person.
In terms of usability, it would be more desirable if the person could be retrieved using more easily available types of queries.

\begin{figure}[t]
\begin{center}
\includegraphics[width=\linewidth]{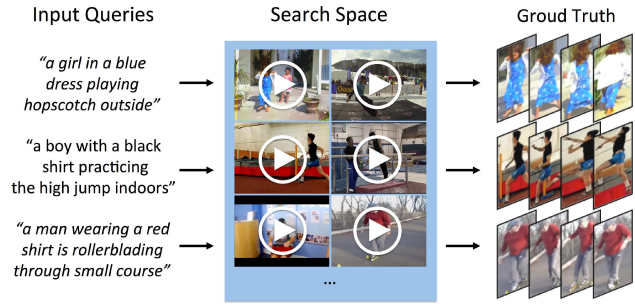}
\end{center}
\caption{Overview of our task. A person is retrieved via a description in the form of a ``tube''.}
\figlab{overview}
\end{figure}

Motivated by this demand, we address the problem of retrieving a person from multiple video clips via a natural language description, which can be easily input and used even if no image of the person is available.
Specifically, we output the target people in the form of tubes (i.e., sequences of bounding boxes), which provide rich information that can be used for various applications, such as video editing.
The overview of our task is shown in \figref{overview}.

To address this problem, we constructed a novel dataset as a platform for evaluating this task.
We first collected \NClip video clips from the ActivityNet dataset \cite{caba2015activitynet}, which contains videos including diverse human actions.
We then chose \NPerson people in those clips and annotated each of them with a bounding box for each second and also with five descriptions.
The total duration of our dataset is 13.7 hours, the number of bounding boxes is 62,627, and the number of descriptions is 30,365.

To solve this problem, we design a model in which we first obtain candidate tubes and then retrieve the top $K$ tubes suitable for a given natural language query.
To obtain candidate tubes, we utilize a spatio-temporal human detection method based on convolutional neural networks (CNNs).
To retrieve tubes via a given query, we utilize a multimodal feature embedding method in combination with rich tube and description representations.
We construct a strong baseline in this task by conducting comprehensive experiments to compare a variety of tube and text representations and multimodal retrieval methods.
Through the experiments we also demonstrate the efficacy of our tube representation and multimodal feature embedding technique.

Moreover, the model trained with our dataset is generic and can be used for various purposes.
To verify the versatility of our model, we apply our model to two important tasks: (1) video-clip retrieval and (2) spatio-temporal action detection, and show our model is also effective in these tasks.


We will make the code, features, and annotations publicly available\footnote{Project page: \url{http://www.mi.t.u-tokyo.ac.jp/projects/person_search/}}.

\section{Related work}
\mysubsection{Spatio-temporal action detection.}
Spatio-temporal action detection is the problem of localizing people performing an action in a set of pre-defined action categories.
A number of approaches to this problem have been developed \cite{tian2013spatiotemporal,wang2014video,chen2015action,gkioxari2015finding,weinzaepfel2015learning,soomro2015action,saha2016deep,peng2016multi}.
The task we address is similar to spatio-temporal action detection in that we output the tube of people suitable for the query, however it is more challenging than spatio-temporal action detection in that a query is given in the form of a nearly-free natural language description.

\mysubsection{Locating objects in images via natural language queries.}
\cite{plummer2015flickr30k,wang2015learning,hu2015natural,mao2015generation,yu2016modeling,nagaraja2016modeling,rohrbach2015grounding,wang2016structured} addressed the problem of localizing objects in images via natural language queries.
The task we address can be seen as a video counterpart of localizing a target in an image based on text queries.
However, our task can be considered more challenging than this task in the following two ways:
(1) we retrieve targets from multiple videos rather than from an image, and
(2) we output the results in the form of sequences of bounding boxes.

\mysubsection{Natural language video search.}
Several works such as \cite{neo2006video, li2007video, snoek2007adding, lin2014visual, siddharth2014seeing} addressed the problem of natural language video search.
Compared to this task in which we return the clips as retrieval results, the task we address can be considered more difficult since we have to return the tubes as results.
Accordingly, the models for our task can be applied to more diverse purposes than those for video search.

\mysubsection{Locating objects in videos via natural language queries.}
The studies most related to this paper are \cite{siddharth2014seeing, lin2014visual}, which addressed the problem of grounding objects described in a given sentence in a given video clip.
\cite{siddharth2014seeing} localized objects using their sentence-tracker, which matches nouns in a given sentence and objects in a given video based on pre-defined rules for visual attributes such as colors and actions.
\cite{siddharth2014seeing} used a dataset that comprises 94 short video clips, containing limited types of actors, objects, and actions, using similar scenes filmed from similar camera angles.
\cite{lin2014visual} proposed a method that first parses a given sentence into a semantic graph, then matches noun nodes in the graph and detected objects using a generalized bipartite matching algorithm in combination with pre-defined scoring functions for visual attributes such as object appearance and motion.
\cite{lin2014visual} used a dataset comprised of 21 videos filmed by an on-vehicle camera, of which the average frame length was 381.
Note that there are three key differences between the problem settings in existing works \cite{siddharth2014seeing, lin2014visual} and our study:\\
(1) \cite{siddharth2014seeing, lin2014visual} focused on localizing one or more nouns included in a given query, whereas we assum that a given query describes just one person and focus on localizing him/her.\\
(2) \cite{siddharth2014seeing, lin2014visual} addressed the problem of localizing targets in a single video clip, whereas we address the problem of retrieving targets from multiple video clips.\\
(3) The videos used in \cite{siddharth2014seeing, lin2014visual} were filmed under limited conditions or in limited environments, whereas the videos used in our work are derived from the ActivityNet dataset \cite{caba2015activitynet}, of which videos are very diverse in terms of backgrounds, camera angles, actors, human actions, and human attributes.\\
It is also worth noting that the methods used in \cite{siddharth2014seeing, lin2014visual} require a pre-defined rule or scoring function for each visual attribute.
Contrarily, our model utilizes a multimodal retrieval method in combination with recently-proposed rich text and visual representations, and can retrieve a target described by a given query without any pre-defined rules or scoring functions for visual attributes.

\section{Dataset}
To tackle the problem of retrieving a person in videos using a natural language query, we created a novel dataset, which consists of videos containing people annotated with bounding boxes every second and with five descriptions
Dataset examples are shown in \figref{dataset}.
We first explain how  our data was collected, and then show the statistics of our dataset.

\begin{figure*}[t]
\begin{center}
\includegraphics[width=\linewidth]{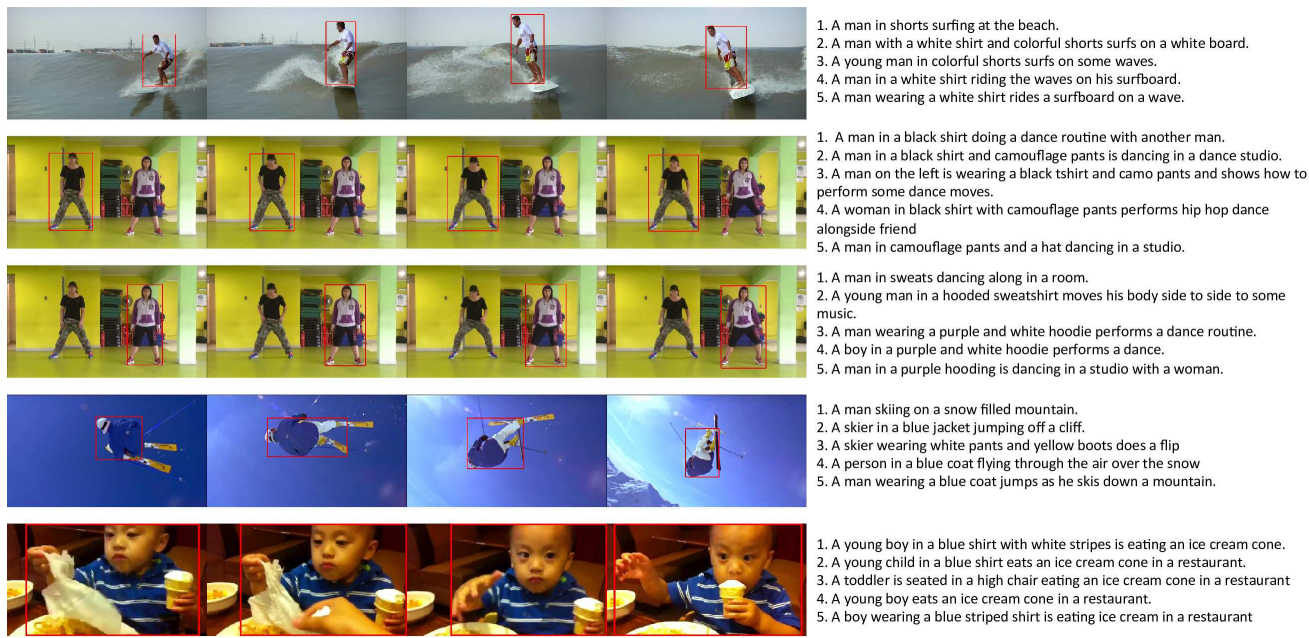}
\end{center}
\caption{Examples of people and annotated descriptions in our dataset. We show only four frames for each of five samples.}
\figlab{dataset}
\end{figure*}

\subsection{Dataset Collection}
\subsubsection{Selection of Video Clips and People}\seclab{chooseclip}
First, we collected short video clips for the dataset.
To diversify visual content in our dataset, we chose clips from videos in the ActivityNet dataset \cite{caba2015activitynet}, which covers a wide range of complex human activities.
We first split each of the activity instances in ActivityNet into shots based on the L1 distance between color-histograms of adjacent frames, then checked whether there are any people who satisfy all of the following conditions in each shot:
(1) the head of the person can be seen,
(2) the person is one of the main people in the shot,
(3) the person exists for more than 5 seconds, and
(4) the person is displaying any characteristic actions, or his or her appearance is characteristic.
If there were any people who satisfied the above conditions in the shot, we chose a period of around 10 seconds from the period in which these people exist in the shot.
To ensure the diversity of the dataset, for training data we chose at most two clips for each video, and for validation and testing data we chose at most one clip for each video.

\subsubsection{Bounding Box Annotation}\seclab{bboxann}
Next, we annotated each person who satisfied the conditions explained in \secref{chooseclip} with a bounding box every second, with the help of in-house annotators.
In case where the target person could not be seen because of occlusion or framing out, we asked them not to annotate the person with a bounding box.
To improve the quality of the annotation, we next checked all the annotated bounding boxes and modified inappropriately ones.

In addition to bounding boxes, we also annotated each clip with a binary label every second, to indicate whether the frame of the timestep includes any people who are not annotated with bounding boxes.
Namely, in the timestep of which the binary label is true, it is guaranteed that there are one or more people not annotated with bounding boxes.
These labels are useful when training human detectors using this dataset since we can remove images that include unlabeled humans using these labels.

\subsubsection{Description Annotation}\seclab{description}
Next, we annotated each person with five natural language descriptions using Amazon Mechanical Turk.
Following \cite{rashtchian2010collecting}, we used a brief qualification test to prevent workers whose English skills were insufficient from joining the task.
For each target person, we asked five workers to describe the person under the following conditions:
(1) the description must contain at least eight words,
(2) the description must contain all the important parts of the person such as their clothes and actions,
(3) the description must be unique to the target person in case multiple people are present, and
(4) the description should also include information about the scene around the person if it can be described.

In cases where multiple people were present, it was necessary to prevent workers from describing people other than the target person.
We solved this problem by using a tool we developed to indicate the target person by repeatedly redrawing a bounding box enclosing him/her on the embedded video element.

To improve the quality of the annotated descriptions, we finally asked workers to correct descriptions that contains spelling or grammatical errors, or does not satisfy any of the above conditions.

\subsection{Dataset Statistics}
Statistics of the training, validation, and testing datasets are included in \tabref{dsstat}.
We collected \NClip clips and annotated \NPerson people in those clips with 30,365 descriptions and 62,627 bounding boxes in total.
The number of unique words is 6,730, the total number of words is 398,397 and the average length of descriptions is 13.1.
We discuss the statistics of our dataset further in the supplementary.

\begin{table}[t]
\begin{center}
\small
\begin{tabular}{l|ccccc}
\hline
 & \#Clip & Duration & \#Person & \#Description & \#Box\\
\hline\hline
train & 4,734 & 732 min. & 5,437 & 27,185 & 55,875 \\
val & 276 & 44 min. & 313 & 1,565 & 3,311 \\
test & 283 & 46 min. & 323 & 1,615 & 3,441 \\
\hline
\end{tabular}
\end{center}
\caption{Dataset statistics. We discuss other statistics further on the supplementary.}
\tablab{dsstat}
\end{table}

\section{Approach}
\subsection{Overview}
We aim to retrieve the tubes suitable for a person described by a given natural language description from a set of video clips.
To this end, inspired by the recent method for object detection \cite{girshick2014rich}, we design a model in which we first propose candidate tubes, rank them using a method for multimodal retrieval based on a given query, and then output the top $K$ scoring tubes.
\figref{system} shows an overview of our method.
For obtaining candidate tubes, we utilize a method for spatio-temporal person detection.
For ranking candidate tubes based on a given query, we extract features from tubes and the text query, and then retrieve tubes suitable for the query based on extracted features using a multimodal retrieval method.

In the following, we present details of the modules.

\begin{figure}[t]
  \begin{center}
    \includegraphics[width=\linewidth]{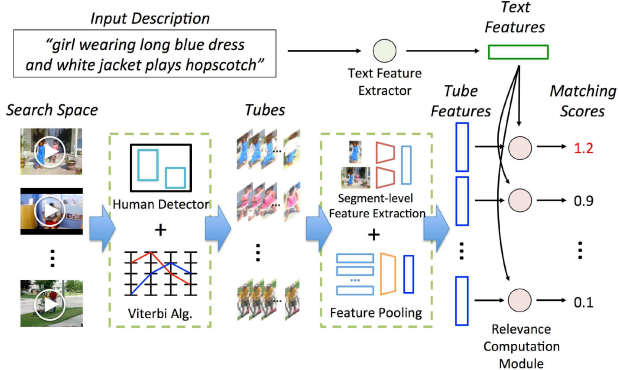}
  \end{center}
  \caption{Diagram of our model. We first propose candidate tubes, and then extract features of them. Given a description, we extract its text features, and then compute the matching score between the description and each tube using extracted features.}
  \figlab{system}
\end{figure}

\subsection{Candidate Tube Proposal}\label{sec:spatiotemporaldet}
Some researchers previously proposed methods \cite{yu2015fast,van2015apt,oneata2014spatio,jain2014action,xiao2016track,marian2015unsupervised} for obtaining candidate tubes of human actions, however, many of these utilize motion information for obtaining candidate tubes and such methods may fail to propose candidate tubes for people who move subtly or infrequently.
Alternatively, we directly obtain candidate tubes by using spatio-temporal human detection.
More specifically, we detect human tubes with a method inspired by the action detection method proposed in \cite{gkioxari2015finding}, in which we first detect humans in each frame, and then obtain candidate tubes by linking the detected bounding boxes.
In the following, we explain the details of those two steps.

\subsubsection{Human Detection in each frame}
First, we detect humans in each frame.
For human detection, we use Faster R-CNN \cite{ren2015faster} as a detector for static images and employ VGG 16-layer net \cite{simonyan2014very} pretrained on the ImageNet dataset \cite{deng2009imagenet} as a base network within Faster R-CNN.
For training and early stopping, we use our training and validation datasets, respectively.
The approximate joint training strategy introduced in \cite{ross2015rcnn} is used to train the model.

\subsubsection{Candidate tube generation}
To create spatio-temporal tubes, we next link up the human detections between frames.
As with \cite{gkioxari2015finding}, we define the linking score $s_{link}(B_t, B_{t+1})$ between the $i_t$-th bounding box $B_{t,i_t}$ at time $t$ and the $i_{t+1}$-th bounding box $B_{t+1,i_{t+1}}$ at time $t+1$ as follows:
\begin{equation}
  \begin{split}
    s_{link}(B_{t,i_t}, B_{t+1,i_{t+1}}) = & s_{det}(B_{t,i_t}) + s_{det}(B_{t+1,i_{t+1}}) + \\
    & \lambda {\rm OV}(B_{t,i_t}, B_{t+1,i_{t+1}}),
  \end{split}
\end{equation}
where $s_{det}(B)$ is the human score of the bounding box $B$ obtained by Faster R-CNN, ${\rm OV}(B_{t,i_t}, B_{t+1,i_{t+1}})$ is the intersection over union between those two bounding boxes, and $\lambda$ is a hyperparameter.

As with \cite{gkioxari2015finding}, we next define the energy function $E(\overline{B})$ of the tube $\overline{B}=\bigl[B_{1, i_1},B_{2, i_2},...,B_{T, i_T}\bigr]$ as follows:
\begin{equation}
  E(\overline{B}) = \frac{1}{T} \sum_{t=1}^{T-1} s_{link}(B_{t, i_t}, B_{t+1, i_{t+1}}) \equlab{detscore}.
\end{equation}
We can obtain the optimal path $\overline{B}^* = \argmax E(\overline{B})$ by using the Viterbi algorithm.
We repeatedly obtain the optimal path by solving \equref{detscore} and removing the bounding boxes included in the obtained path until at least one set of bounding boxes included in the same timestep is empty.
We finally choose the top $N_c$ scoring tubes from those obtained from all the clips in the target dataset, and use them as the candidate tubes.

\subsection{Ranking tubes via a description}\seclab{tubesearch}
We next explain the approach to ranking tubes given a natural language description.
Inspired by the state-of-the-art method \cite{wang2015learning} in the image-sentence retrieval task, we employ the approach in which we project features of both modalities into the common space and retrieve candidate tubes in the space.
We first extract the tube features from candidate tubes and the text features from a given description.
We then project the features of those tubes and the given description into the common space, and rank the candidate tubes based on the distance to the description in that space.

In the following, we explain the tube and description features and the multimodal retrieval methods we use.

\subsubsection{Tube features}

\begin{figure}[t]
  \begin{center}
    \includegraphics[width=\linewidth]{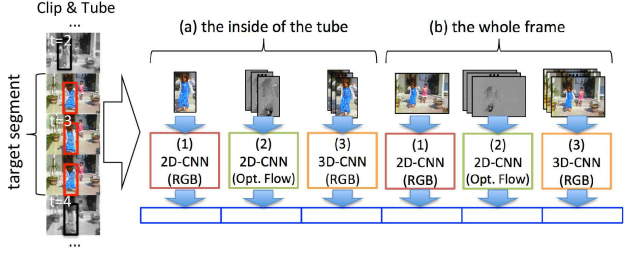}
  \end{center}
  \caption{Diagram of extracting the segment-level features. For each tube, we compute and concatenate the above six types of features every second.}
  \figlab{shortterm}
\end{figure}

To extract tube features from a given tube, we first extract the segment-level features once per second, and then construct the tube's feature vector by aggregating those features.

\mysubsection{Segment-level features.}
As shown in \figref{shortterm}, we extract:\\
(1) the CNN features extracted from the RGB images,\\
(2) the CNN features extracted from the sequences of the optical flow maps, and\\
(3) the 3D-CNN (or C3D) features \cite{tran2014learning} extracted from the sequences of RGB images\\
from (a) the inside of tubes and (b) the whole frames for each.
We then construct one feature vector by concatenating all of six types of the above features and use it as the segment-level features.

To extract the CNN features from the RGB images, we use the outputs from the fc7 layer in the VGG-16 layer net pretrained on ImageNet.
To extract the CNN features from the optical flow images, we use the outputs from the fc7 layer in the VGG-16 layer net first pretrained on ImageNet and then re-trained on the first split of the UCF101 dataset \cite{soomro2012ucf101}.
We used the model provided by the CUHK team on the web\footnote{https://github.com/yjxiong/caffe} \cite{wang2015towards}.
Following the CUHK team, we extract the optical flow maps by using the TV-L1 optical flow algorithm \cite{zach2007duality} and input the optical flow maps corresponding to the five frames preceding the target timestep and the five frames following the target timestep into the model.
To extract the C3D features, we use the outputs from the fc7 layer in the model pretrained on the Sports-1M dataset \cite{KarpathyCVPR14}\footnote{http://www.cs.dartmouth.edu/~dutran/c3d/}. We input the RGB images corresponding to the frame at the target timestep, the 7 frames preceding it, and the 8 frames following it into the model.

The proposed segment-level features have the following ideal properties:
\begin{itemize}
  \setlength{\parskip}{-2mm}
  \setlength{\itemsep}{2mm}
\item Thanks to the features extracted from (a), the features would include information for the inside of the tube.
\item Thanks to the features extracted from (b), the features would include context information.
\item Thanks to (1) and (3), the features would include appearance information.
\item Thanks to (2) and (3), the features would include motion information.
\item Thanks to (2) and (3), the features would include multi-frames' information-integrated information.
\end{itemize}

\mysubsection{Feature aggregation.}
We next aggregate the set of the segment-level features.
We compared three aggregation strategies (mean/max pooling and encoding with an RNN) and combinations of them, and found that even simple mean pooling works well.
Therefore, in all experiments in \secref{exp}, we use mean pooling for feature aggregation.
We show the results of the comparison of the aggregation strategies in the supplementary.

\subsubsection{Description features}
We compared the following three methods for extracting text features and combinations of them: the Fisher Vectors (FVs) based on a hybrid Gaussian-Laplacian mixture model (HGLMM) \cite{klein2015associating}, skip-thought vectors \cite{kiros2015skip} and one method based on an RNN trained from scratch.
We found that using the FVs based on HGLMM as is works best.
Therefore, in all experiments in \secref{exp}, we use the FVs based on HGLMM as text features.
We explain the details of other two text features and the results of comparison of three text features in the supplementary.

To compute the FVs based on HGLMM, we first apply Independent Component Analysis (ICA) for 300-dimensional word2vec vectors\footnote{https://code.google.com/archive/p/word2vec/} \cite{mikolov2013distributed} and train an HGLMM with 30 centers using ICA-applied word vectors.
Next, we compute the FVs of descriptions using the learned HGLMM and apply power and L2 normalizations to them.
We also apply PCA, as we obtained higher retrieval accuracy than when using the original FVs.
We set the number of dimensions after reduction to 1,000 based on the results obtained in the validation dataset.

\subsubsection{Multimodal Retrieval Methods}
In this study, we compare the three following methods for multimodal retrieval:

\mysubsection{Canonical Correlation Analysis.}
The first method is Canonical Correlation Analysis (CCA), which is much simpler than complex neural networks yet achieves high accuracy in various retrieval tasks\cite{klein2015associating,plummer2015flickr30k}.
When using CCA, we retrieve tubes using the matching score $S_{CCA}(i, j)$ between the $i$-th tube features $f^t_i$ and the $j$-th description features $f^d_j$, which is computed as follows:
\begin{equation}
  S_{CCA}(i, j) = {\rm cos}({\rm diag}(r)\cdot W^t\cdot f^t_i, {\rm diag}(r)\cdot W^d\cdot f^d_j) \equlab{tubescore},
\end{equation}
where ${\rm cos}(x_1, x_2)$ is the cosine similarity between $x_1$ and $x_2$, ${\rm diag}(r)$ is a diagonal matrix whose diagonal elements are eigenvalues learned by CCA, and $W^t$ and $W^d$ are the projection matrices of the tube and the description features into the common latent space, respectively.

\mysubsection{Deep Structure-Preserving Embedding.}
Deep Structure-Preserving Embedding (DSPE)\cite{wang2015learning} is the state-of-the-art method for image-sentence retrieval.
DSPE embeds features of two different modalities into the common space using two neural networks trained with the following loss function:
\begin{equation}
  \begin{split}
    L_{DSPE} = & \sum_{i,j,k}max\left[0, m + {\rm d}(x_i, y_j) - {\rm d}(x_i, y_k) \right]\\
    & + \alpha_1 \sum_{i',j',k'}max\left[0, m + {\rm d}(x_j', y_i') - {\rm d}(x_k', y_i')\right]\\
    & + \alpha_2 \sum_{i,j,k}max\left[0, m + {\rm d}(x_i, x_j) - {\rm d}(x_i, x_k)\right]\\
    & + \alpha_3 \sum_{i',j',k'}max\left[0, m + {\rm d}(y_i', y_j') - {\rm d}(y_i', y_k')\right]
  \end{split}, \equlab{dspeloss}
\end{equation}
where {\rm d} is the distance function between two embedded features, $\alpha_i(i=1,2,3)$ and $m>0$ are hyper-parameters, $(x_i, y_j)$ and $(x_j', y_i')$ are positive pairs, $(x_i, y_k)$ and $(x_k', y_i')$ are negative pairs, $(x_i, x_j)$ and $(y_i', y_j')$ are pairs of examples that are of the same modality and correspond to the same sample of the other modality, and $(x_i, x_k)$ and $(y_i', y_k')$ are pairs that do not.
When using DSPE, we retrieve tubes using the matching score $S_{DSPE}(i, j)=-d(f^t_i, f^d_j)$.
Following \cite{wang2015learning}, we use the network architecture shown in \tabref{dspe}, and employ the Euclidian distance as the distance function ${\rm d}$.

\begin{table}[t]
  \begin{center}
    \small
    \begin{tabular}{lc}
      \hline
      {\bf Layer type} & {\bf \# Filters} \\
      \hline
      \hline
      (Input) & \\
      Inner Product & 2048 \\
      ReLU & \\
      Inner Product & 512 \\
      Batch Normalization \cite{ioffe2015batch} & \\
      ReLU & \\
      L2 Normalization & \\
      (Output) & \\
      \hline
    \end{tabular}
  \end{center}
  \caption{The network architecture of DSPE used in this paper for each module. We also use Dropout \cite{srivastava2014dropout} ($p=0.5$) after the first Inner Product layers.}
  \tablab{dspe}
\end{table}

\mysubsection{Modification of DSPE.}
Although DSPE is an effective retrieval method as is, to further improve accuracy we propose to add a simple yet effective term to the DSPE's loss.
In addition to the original loss, which encourages the distance between positive pairs to be smaller than the distances between negative pairs, we propose to additionally use the summation of the negative distance between positive pairs, which directly encourage the model to reduce the distance between positive pairs.
The proposed loss is as follows:
\begin{equation}
  L_{Proposed} = L_{DSPE} + \alpha_4 \sum_{i,j}{\rm d}(x_i, y_j), \equlab{propall}
\end{equation}
where $\alpha_4$ is a hyper-parameter. We use the same network architecture and distance metric as DSPE.
We refer to the embeddings trained with this loss as DSPE++, and use it as the third choice.

\section{Experiments}\seclab{exp}
We conduct comprehensive experiments to compare (1) the number of candidate tubes to use, (2) the tube representations, (3) the multimodal retrieval methods, (4) the strategies for tube feature aggregation, and (5) the text representations.
We discuss the former three in this section, and the latter two in the supplementary.

\begin{figure}[t]
  \begin{center}
    \includegraphics[width=\linewidth]{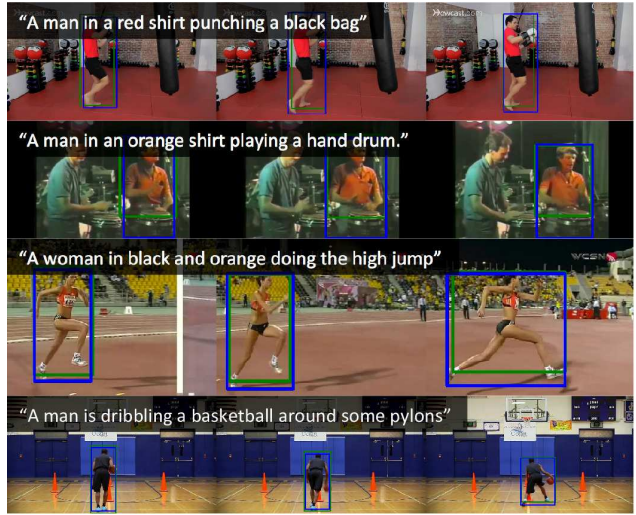}
  \end{center}
  \caption{Correctly retrieved examples. The blue and green bounding boxes are the ground truth and the top 1 retrieved result, respectively. Note that the search space consists of 283 video clips with a total duration of 46 minutes.}
  \figlab{good}
\end{figure}

\begin{table*}[t]
  \begin{center}
    \small
    \begin{tabular}{cccccc|ccc}
      \multicolumn{2}{c}{VGG16 (RGB)} & \multicolumn{2}{c}{VGG16 (Opt. Flow)} & \multicolumn{2}{c|}{C3D} & & & \\
      Tube & Context & Tube & Context & Tube & Context & R@1 & R@5 & R@10 \\
      \hline
      \hline
      O & - & - & - & - & - & 0.276 & 0.572 & 0.693 \\
      O & O & - & - & - & - & 0.279 & 0.623 & 0.751 \\
      O & O & O & O & - & - & 0.285 & 0.620 & 0.760 \\
      O & O & - & - & O & O & {\bf 0.300} & 0.654 & 0.788 \\
      O & O & O & O & O & O & {\bf 0.300} & {\bf 0.658} & {\bf 0.792} \\
      \hline
    \end{tabular}
  \end{center}
  \caption{Performance comparison of tube representations. In these experiments, CCA is used for retrieval.}
  \tablab{imfeat}
\end{table*}

\begin{table}[t]
  \begin{center}
    \small
    \begin{tabular}{c|ccc}
      Multimodal Retrieval Method & R@1 & R@5 & R@10 \\
      \hline
      \hline
      CCA & 0.300 & 0.658 & 0.792 \\
      DSPE & 0.347 & 0.687 & 0.783 \\
      DSPE++ (Proposed) & {\bf 0.357} & {\bf 0.702} & {\bf 0.795} \\
      \hline
    \end{tabular}
  \end{center}
  \caption{Performance comparison of multimodal retrieval methods.}
  \tablab{mrm}
\end{table}

\subsection{Settings}
For this task, we address the problem of retrieving a tube for a person described by each query from video clips in the test dataset, of which the total duration is approximately 46 minutes.
We use 1,615 descriptions from the test dataset as the queries.

\mysubsection{Evaluation Metric.}
The retrieval task can be evaluated using R@K, that is, the recall rate of the ground truth included in the top K candidates retrieved using a given query.
To use the R@K as metrics of our tasks, we must define the conditions to judge whether a retrieved instance can be considered as a true positive of the ground truth.
We compute R@K by assuming $dt_{tube}$ is the true positive of $gt_{tube}$ in the case that the following localization score $S_{loc}(gt_{tube}, dt_{tube})$ is over $0.5$:
\begin{equation}
  S_{loc}(gt_{tube}, dt_{tube}) =\frac{1}{|\Gamma|}\sum_{f\in\Gamma}{\rm OV}(gt_{tube,f}, dt_{tube,f}),
\end{equation}
where $\Gamma$ is the intersection of the set of frames to be annotated with bounding boxes and the set of frames in which $gt_{tube}$ or $dt_{tube}$ has any bounding box.

\mysubsection{Training models.}
For training the retrieval methods, we use 27,185 tube-description pairs in our training dataset.
We show the details of training and hyper-parameters for them in the supplementary.

\subsection{Results}

\mysubsection{Influence of the number of candidate tubes.}
Our model first chooses the top $N_c$ scoring tubes from tubes obtained by the candidate proposal module, and then ranks those based on a given description.
We first conducted experiments to investigate the influence of the number of candidate tubes $N_c$.
In this experiment, we used CCA for retrieval and computed the recall rates by setting $N_c=100,150,...,700$.


\begin{figure}[t]
  \begin{center}
    \includegraphics[width=0.5\linewidth] {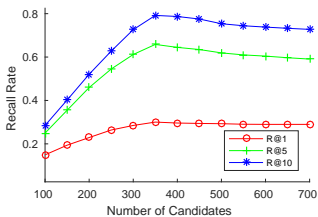}
  \end{center}
  \vspace{-3mm}
  \caption{Retrieval accuracy of our method on val dataset.}
  \figlab{ncands}
\end{figure}

The results for the validation dataset are shown in \figref{ncands}.
Considering all metrics, our method achieves the highest retrieval accuracy for $N_{c}=350$, and retrieval accuracy decreases as the number of candidates becomes smaller or larger than $N_{c}=350$.
This decrease occurs because if $N_c$ is too small, the probability that the chosen tubes include any true positives of the ground truth is too low, and this reduces retrieval accuracy even further. Contrarily, if $N_c$ is too large, the amount of noise in the chosen tubes increases and the retrieval accuracy again declines.

Because of this result and the fact that the size of the testing dataset is almost the same as that of the validation dataset, we fix $N_{c}=350$ in the following experiments.

\mysubsection{Comparison of tube representations.}
Here we discuss the experiments conducted to compare tube representations.
In these experiments, we investigate the efficacy of the context information and three types of features explained in \secref{tubesearch}.
CCA is used for retrieval.

We show the comparison of the tube representations in \tabref{imfeat}.
In \tabref{imfeat}, the cases in which the CNN features from the optical flow maps, the C3D features, or the features extracted from whole frames are used achieve higher accuracy than the cases in which these features are not used. The case in which the full range of features are used achieves the highest accuracy.
These results suggest the importance of considering not only appearance, but motion, context, and multi-frame information simultaneously in this task.

\mysubsection{Comparison of multimodal retrieval methods.}
Next, we discuss the experiments to compare the multimodal retrieval methods.

The results are presented in \tabref{mrm}.
We can see that while CCA and DSPE are comparable, DSPE++ outperforms them in all metrics.
These results suggest the effectiveness of the proposed loss in this task\footnote{We also conducted experiments to compare DSPE and DSPE++ in the image-sentence retrieval task. We discuss them in the supplementary.}.

\mysubsection{Qualitative results.}
The correctly retrieved examples are shown in \figref{good}.
We stress that the search space consists of 283 video clips with a total duration of 46 minutes.
These examples imply that our method can find a plausible person tube based on a complex query containing various types of information such as clothes and actions.
The additional correctly retrieved examples and randomly chosen examples are also shown in the supplementary.

\subsection{Applying our model to other tasks}
Although our model is designed to retrieve a person tube based on a natural language description, our model is generic and it can be used for various purposes.
In this section, we verify its versatility by applying our trained model to two other important tasks: the clip retrieval task and the spatio-temporal action detection task.

\subsubsection{Clip Retrieval}
For certain applications, such as video search, it is sufficient that the model just returns the clip containing the target person rather than the tube of the person.
Therefore, we conduct experiments of the clip retrieval task to investigate whether using our model is also effective for this task.

\mysubsection{Setup.}
For this task, we address the problem of retrieving a clip including a person described by each query from the set of video clips in the test dataset.
We used 1,615 desriptions in the test dataset as the queries.
We used R@K as the evaluation metric.

\mysubsection{Baseline.}
As a baseline, we employ a method that directly learns the common space between the description features and clip features using DSPE++.
As clip features, we use the features obtained by concatenating the temporal means of the three types of features (i.e., the CNN features from RGB images, the CNN features from optical flow maps, and the C3D features), extracted from the whole frames.
In the training phase, we consider the description and the clip containing the person described by the description as one training pair and train DSPE++ with 27,185 clip-description pairs.

\mysubsection{Retrieval with our model.}
We propose to define the matching score between a description and a clip by the maximum of the score between the description and a candidate tube in the clip computed by our trained model.
More specifically, we employ the following matching score between the $i$-th description and the $j$-th clip for retrieving a clip:
\begin{equation}
  S_{proposed}^*(i, j) = \max_{k=k_{i,1},...,k_{i,N_i}} S(i, k),
\end{equation}
where $S(i, j)$ is the relevance score between $i$-th description and $k$-th tube computed by our model, $\bigl[k_{j,1},...,k_{j,N_i}\bigr]$ are indices of the candidate tubes in the $j$-th clip and $N_j$ is the number of candidate tubes in the clip.

\mysubsection{Results.}
\tabref{cliplevel} contains the results.
All metrics indicated that higher retrieval accuracy can be achieved by our method than by using the baseline method.
This result suggests that our model is also useful for clip retrieval tasks.

\begin{table}[t]
  \begin{center}
    \small
    \begin{tabular}{l|ccc}
      & R@1 & R@5 & R@10 \\
      \hline
      \hline
      Baseline & 0.373 & 0.749 & 0.879 \\
      Proposed & {\bf 0.415} & {\bf 0.806} & {\bf 0.893} \\
      \hline
    \end{tabular}
  \end{center}
  \caption{Retrieval accuracy for the clip-level person retrieval task. Comparison of the baseline method and the method that leverages our method for spatio-temporal person retrieval task.}
  \tablab{cliplevel}
\end{table}

\subsubsection{Action Detection}
\begin{figure}[t]
  \begin{center}
    \includegraphics[width=\linewidth]{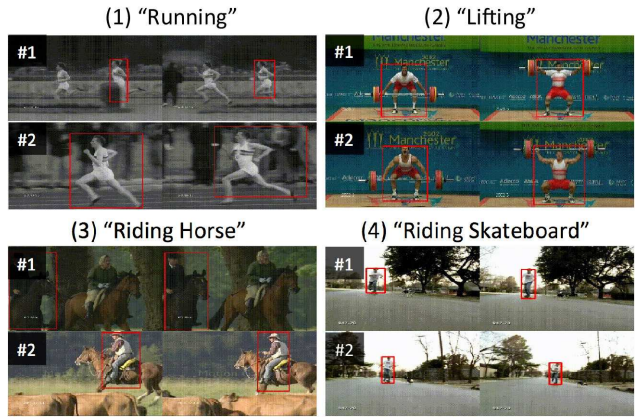}
  \end{center}
  \caption{Top 2 tubes retrieved using each of four queries. Two frames are shown for each tube.}
  \figlab{simple}
\end{figure}

Spatio-temporal action detection is an important task in video analysis.
Although our model is designed to retrieve a person based on a natural language description, it can be also used for this task using an action category name as a query.
To verify this, we conduct an experiment to test the efficacy of our model as a solution for spatio-temporal action detection on the UCF Sports dataset \cite{rodriguez2008action}, which consists of 150 video clips with 10 action classes.

\figref{simple} shows the examples retrieved from the dataset using each of the following four queries: ``Running'', ``Lifting'', ``Riding Horse'' and ``Riding Skateboard''.
Even though our model is trained without the data in the UCF Sports dataset, these action instances can be correctly detected by just inputting the action name to our trained model.
These results indicate that our model can be also used as a human-action detector.
The quantitative results are also presented in the supplementary.

\section{Conclusion}
In this paper, we have addressed spatio-temporal person retrieval from multiple videos using natural language queries.
We present a dataset of videos containing people annotated with bounding boxes and descriptions.
We also present a model that combines methods for spatio-temporal human detection and multimodal retrieval.
We optimize the model by conducting comprehensive experiments to compare tube and text representations and multimodal retrieval methods, and construct a strong baseline on this task.
Finally, we validate our model's versatility by conducting experiments on two important additional tasks.

\section{Acknowledgments}
This work was supported by JST CREST Grant Number JPMJCR1403, Japan.
This work was also supported by the Ministry of Education, Culture, Sports, Science and Technology (MEXT) as ``Seminal Issue on Post-K Computer''.

{\small
\bibliographystyle{ieee}
\bibliography{paper_final}
}

\end{document}